\documentclass[11pt,letterpaper]{article}
\usepackage{etex}

\usepackage{emnlp2016}
\usepackage{times}
\usepackage{latexsym}

\usepackage{graphicx}
\usepackage{amsfonts}

\usepackage[hang,flushmargin]{footmisc} 

\usepackage{changepage}

\usepackage{nohyperref}
\hypersetup{
  colorlinks=true,
  citecolor=black,
  linkcolor=black,
  urlcolor=black,
  }

\usepackage[table]{xcolor}
\usepackage{array}
\usepackage{url}

\usepackage{enumitem}
\setlist{nosep}

\usepackage{multirow}
\usepackage{tabularx}
\usepackage{multicol}
\usepackage{multirow}
\usepackage{diagbox}
\usepackage{amsmath}
\usepackage{mathtools}
\usepackage{bm}
\usepackage{booktabs}
\usepackage{graphicx}
\usepackage{caption}
\usepackage{floatrow}
\usepackage[position=top]{subfig}
\restylefloat{table}
\usepackage[font=footnotesize,labelfont=bf]{caption}
\usepackage{enumitem}
\usepackage{makecell}
\newcolumntype{x}[1]{>{\centering\arraybackslash}p{#1}}

\usepackage{multirow,bigdelim,dcolumn,booktabs}

\usepackage{arydshln}

\usepackage{tikz}
\usetikzlibrary{matrix,decorations.pathreplacing,calc}

\pgfkeys{tikz/mymatrixenv/.style={decoration=brace,every left delimiter/.style={xshift=3pt},every right delimiter/.style={xshift=-3pt}}}
\pgfkeys{tikz/mymatrix/.style={matrix of math nodes,left delimiter=[,right delimiter={]},inner sep=2pt,column sep=1em,row sep=0.5em,nodes={inner sep=0pt}}}
\pgfkeys{tikz/mymatrixbrace/.style={decorate,thick}}

\emnlpfinalcopy

\def\emnlppaperid{***}

\title{A naive but challenging baseline for mapping distributional to model-theoretic semantic spaces}
\title{Mapping distributional to model-theoretic semantic spaces: a baseline}

\author{
Franck Dernoncourt \\
MIT\\
\texttt{francky@mit.edu}
}

\date{}

\begin{document}

\maketitle
 \vspace{-0.5cm}
\begin{abstract}
 \vspace{-0.2cm}
 Word embeddings have been shown to be useful across state-of-the-art systems in many
natural language processing tasks, ranging from question answering systems to dependency parsing. \cite{herbelot2015building} explored word embeddings and their utility for modeling language semantics. In particular, they presented an approach to automatically map a standard distributional semantic space onto a set-theoretic model using partial least squares regression. We show in this paper that a simple baseline achieves a +51\% relative improvement compared to their model on one of the two datasets they used, and yields competitive results on the second dataset.
\end{abstract}

 \vspace{-0.4cm}

\section{Introduction}

Word embeddings are one of the main components in many state-of-the-art systems for natural language processing (NLP), such as language modeling~\cite{mikolov2010recurrent}, text classification~\cite{socher2013recursive,kim2014convolutional,blunsom2014convolutional,lee2016sequential}, question answering~\cite{weston2015towards,wang-nyberg:2015:ACL-IJCNLP}, machine translation~\cite{bahdanau2014neural,tamura2014recurrent,sundermeyer2014translation}, as well as named entity recognition~\cite{collobert2011natural,dernoncourt2016identification,lample2016neural,labeau-loser-allauzen:2015:EMNLP}.

Word embeddings can be pre-trained using large unlabeled datasets typically based on token co-occurrences~\cite{mikolov2013distributed,collobert2011natural,pennington2014glove}. They can also be jointly learned with the task. 

Understanding what information word embeddings contain is subsequently of high interest. \cite{herbelot2015building} investigated a method to map word embeddings to formal semantics, which is the center of interest of this paper. Specifically, given a feature and a word vector of a concept, they tried to automatically find how often the given concept has the given feature. For example, the concept \textit{yam} is always a \textit{vegetable}, the concept \textit{cat} has a coat most of the time, the concept \textit{plug} has sometimes 3 prongs, and the concept \textit{dog} never has wings. 

The method they used was based on partial least squares regression (PLSR). We propose a simple baseline that outperforms their model.

\section{Task}

In this section, we summarize the task presented in~\cite{herbelot2015building}.
The following is an example of a concept along with some of its features, as formatted in one of the two datasets used to evaluate the model: 

\vspace{-0.2cm}
 \begin{adjustwidth}{-0.1cm}{}
\begin{tabular}{ccccc}
 &  &  &  & \tabularnewline
yam & a\_vegetable & all & all & all\tabularnewline
yam & eaten\_by\_cooking & all & most & most\tabularnewline
yam & grows\_in\_the\_ground & all & all & all\tabularnewline
yam & is\_edible & all & most & all\tabularnewline
yam & is\_orange & some & most & most\tabularnewline
yam & like\_a\_potato & all & all & all\tabularnewline
 &  &  &  & \tabularnewline
\end{tabular}
 \end{adjustwidth}
\vspace{-0.2cm}

The concept \textit{yam} has six features (\textit{a\_vegetable}, \textit{eaten\_by\_cooking}, \textit{grows\_in\-\_the\_ground}, \textit{is\_edible}, \textit{is\_orange}, and \textit{like\_a\_potato}). Each feature in this dataset is annotated by three different humans. The annotation is a quantifier that reflects how frequently the concept has a feature. Five quantifiers are used: \textit{no}, \textit{few}, \textit{some}, \textit{most}, and \textit{all}. In this example, the concept \textit{yam} has been annotated as \textit{some}, \textit{most} and \textit{most} for the feature \textit{is\_orange}.

Each of the five quantifiers is converted into a numerical format with the following (somehow arbitrary) mapping: $\text{no} \mapsto 0$; $\text{few} \mapsto 0.05$; $\text{some} \mapsto 0.35$; $\text{most} \mapsto 0.95$; $\text{all} \mapsto 1$. The value is averaged over the three annotators. Using this mapping, we can map a concept into a ``model-theoretic vector''
(also called feature vector). If a feature has not been annotated for a concept, then the element in the model-theoretic vector corresponding to the feature will have value $0$. As a result, any element of a model-theoretic vector that has value 0 may correspond to a feature that has either been annotated as \textit{no} by the three annotators, or not been annotated (presumed \textit{no}). Given that there can be many features and it is possible that only some of them are annotated for each concept, the model-theoretic vector may be quite sparse. 

In the yam example, if we only included features annotated with yam, the model-theoretic vector would be as follows: 
\vspace{-0.0cm}
\renewcommand{\arraystretch}{1.5}
$$\begin{bmatrix}
\frac{\text{all}+\text{all}+\text{all}}{3}\\ \frac{\text{all}+\text{most}+\text{most}}{3}\\\frac{\text{all}+\text{all}+\text{all}}{3}\\\frac{\text{all}+\text{most}+\text{all}}{3}\\\frac{\text{some}+\text{most}+\text{most}}{3}\\\frac{1+1+1}{3} 
\end{bmatrix} = \begin{bmatrix}
\frac{1+1+1}{3}\\ \frac{1+0.95+0.95}{3}\\\frac{1+1+1}{3}\\\frac{1+0.95+1}{3}\\\frac{0.35+0.95+0.95}{3}\\\frac{1+1+1}{3} 
\end{bmatrix} \approx 
\begin{bmatrix}1\\0.967\\1\\0.983\\0.75\\1
\end{bmatrix} $$

The additional coordinates corresponding to all the remaining features would be zero. Each concept word will have a vector of the same dimension (number of unique features) in the same dataset. The coordinates mean the same from one concept to another. For example, the feature \texttt{is\_vegetable} appears in the same coordinate position in all the vectors. 

\section{Datasets}

Two datasets are used:
\begin{itemize}
\item The Animal Dataset (AD) \cite{herbelottext} contains 73 concepts and 54 features. All concepts are animals, and for each concept all features are annotated by 1 human annotator. There are 3942 annotated pairs of concept-feature ($73*54=3942$). The dimension of the model-theoretic vectors will therefore be 54.
\item TheMcRae norms (QMR) \cite{mcrae2005semantic} contains 541 concepts covering living and non-living entities
(e.g., alligator, chair, accordion), as well as 2201 features. One concept is annotated with 11.4 features on average by 3 human annotators. There are 6187 annotated pairs of concept-feature ($541*11.4\approx 6187$). The dimension of the model-theoretic vectors will therefore be 2201, and each model-theoretic vector will have on average $2201-11.4=2189.6$ elements set to $0$ due to unannotated features.
\end{itemize}

\begin{figure*}[htb!]
\centering
\includegraphics[totalheight=9cm]{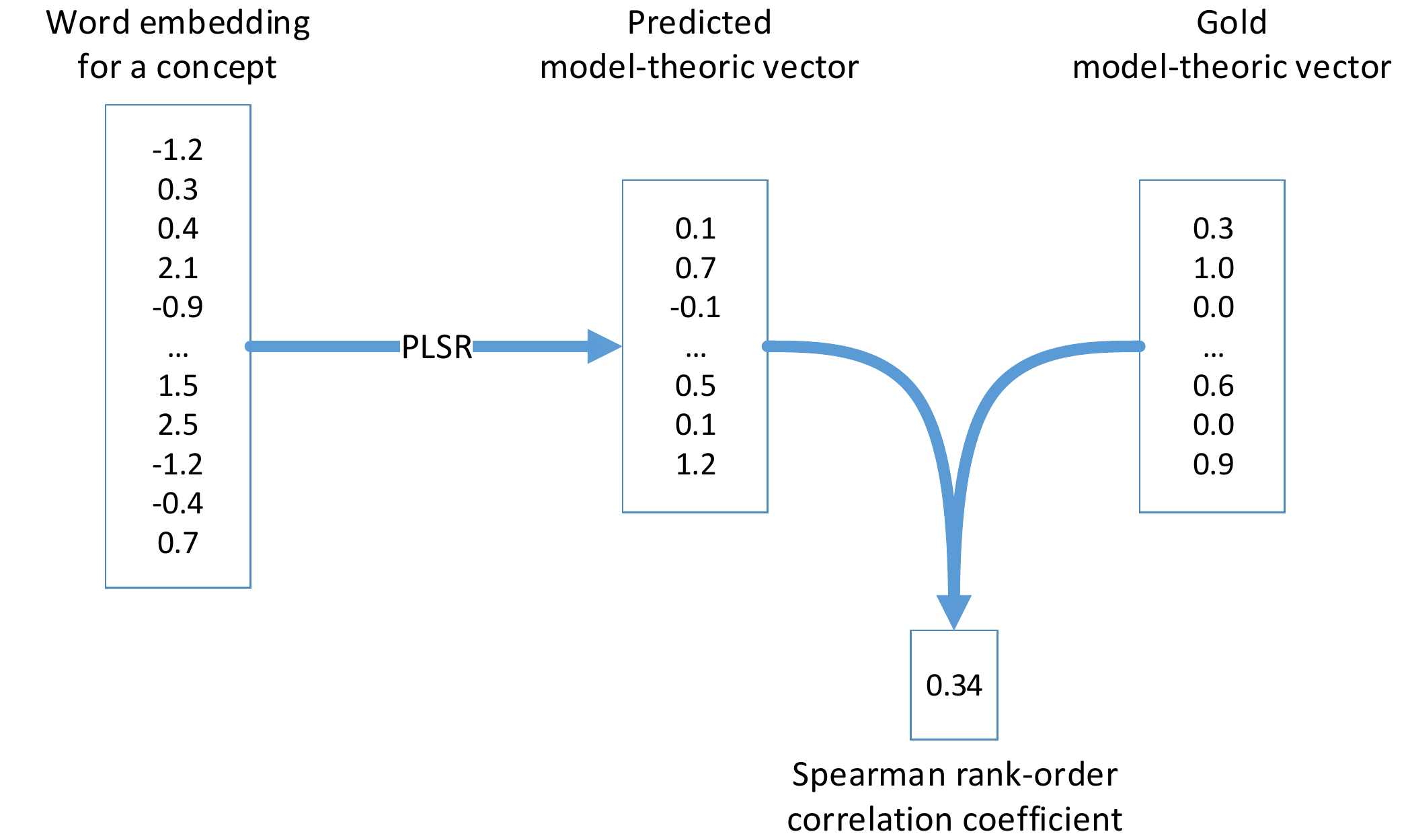}
\caption{Overview of \protect\cite{herbelot2015building}'s system. The word embedding of a concept is transformed to a model-theoric vector via a PLSR. The quality of the predicted model-theoric vector is assessed with the Spearman rank-order correlation coefficient between the predictions and the gold model-theoretic vectors. Note that some of the elements that equal 0 in the gold model-theoretic vector may correspond to features that are not annotated for the concept. Such features are omitted when evaluating the Spearman rank-order correlation coefficient. Also, the dimension of the model-theoretic vectors could be larger or smaller than the dimension of the word embedding. Since the word embeddings we use have 300 dimensions, the model-theoretic vectors will be smaller than the word embeddings in the AD dataset, and larger in the QMR dataset.}
\label{fig:pipeline}
\end{figure*}

\section{Model}

In the previous section, we have seen how to convert a concept into a model-theoretic vector based on human annotations. The goal of \cite{herbelot2015building} is to analyze whether there exists a transformation from the word embedding of a concept to its model-theoretic vector, the gold standard being the human annotations. The word embeddings are taken from the word embeddings pre-trained with word2vec \textit{GoogleNews-vectors-negative300}\footnote{\url{https://code.google.com/p/word2vec/}} (300 dimensions), which were trained on part of the Google News dataset, consisting of approximately 100 billion words.

The transformation used in \cite{herbelot2015building} is based on Partial Least Squares Regression (PLSR).
The PLSR is fitted on the training set: the inputs are the word embeddings for each concept, and the outputs are the model-theoretic vectors for each concept. 

To assess the quality of the predictions, the Spearman rank-order correlation coefficient is computed between the predictions and the gold model-theoretic vectors, ignoring all features for which a concept has not been annotated. The idea is that some of the features might be present but not given as options during annotation. The method should therefore not be penalized for not suggesting them. Figure~\ref{fig:pipeline} illustrates the model.

\section{Experiments}

We compare \cite{herbelot2015building}'s model (PLSR + word2vec) against three baselines: random vectors, mode, and nearest neighbor. 

\begin{itemize}
\item \textit{Mode}: A predictor that outputs, for each feature, the most common feature value (i.e., the mode) in the training set. For example, if a feature is annotated as \textit{all} for most concepts, then the predictor will always output \textit{all} for this feature. When finding the most common value of a feature, we ignore all the concepts for which the feature is not annotated. The resulting predictor does not take any concept into account when making a prediction. Indeed, the predicted values are always the same, regardless of the concept. If a feature has the same value for most concepts, the predictor may perform reasonably well.
\item \textit{Nearest neighbor (NN)}: A predictor that outputs for any concept the model-theoretic vector from the training set corresponding to the most similar concept in the training set. Similarity is based on the cosine similarity of the word vectors. This is a simple nearest neighbor predictor.
\item \textit{Random vectors}: \cite{herbelot2015building} used pre-trained word embeddings as input to the PLSR, we instead simply use random vectors of same dimension (300, continuous uniform distribution between 0 and 1).
\end{itemize}

We also apply retrofitting~\cite{faruqui2014retrofitting} on the word embeddings in order to leverage relational information from semantic lexicons by encouraging linked words to have similar vector representations. Using~\cite{faruqui2014retrofitting}'s retrofitting tool\footnote{\url{https://github.com/mfaruqui/retrofitting}}, we retrofit the word embeddings (\textit{GoogleNews-vectors-negative300}) on each of the 4 datasets present in the retrofitting tool (\textit{framenet}, \textit{ppdb-xl}, \textit{wordnet-synonyms+}, and \textit{wordnet-synonyms}.

\newcolumntype{L}{>{$}l<{$}}
\newcolumntype{C}{>{$}c<{$}}
\newcolumntype{R}{>{$}r<{$}}
\newcommand{\nm}[1]{\textnormal{#1}}

\begin{table*}[htb!]
\centering
\begin{tabular}{LCCCCCC}
\toprule
\multicolumn{1}{l}{} &
\multicolumn{3}{c}{AD}    &
\multicolumn{3}{c}{QMR}    \\ 
\cmidrule(lr){2-4}
\cmidrule(lr){5-7}

&
\multicolumn{1}{c}{Min} &
\multicolumn{1}{c}{Average}     &
\multicolumn{1}{c}{Max}     &
\multicolumn{1}{c}{Min} &
\multicolumn{1}{c}{Average}     &
\multicolumn{1}{c}{Max}     \\
\midrule

\text{PLSR + word2vec}		& 0.435	& 0.572 & 0.713	& 0.244	& 0.332 & 0.407 \\
\text{PLSR + word2vec + framenet}	& 0.423	& 0.577	& 0.710	& 0.236	& 0.331	& 0.410 \\
\text{PLSR + word2vec + ppdb}	& \textbf{0.455}	& 0.583	& 0.688 & 0.247	& 0.332	& 0.421 \\
\text{PLSR + word2vec + wordnet}	& 0.429	& 0.583	& 0.713 & 0.252	& 0.339	& 0.444 \\
\text{PLSR + word2vec + wordnet+}	& 0.453	& \textbf{0.604}	& 0.724 & 0.261	& 0.344	& 0.428 \\
\text{PLSR + random vectors} & 0.253	& 0.419	& 0.550 & -0.017	& 0.087	& 0.178 \\
\text{NN + word2vec} & 0.338	& 0.524	& 0.751 & 0.109	& 0.215	& 0.291 \\
\text{NN + word2vec + framenet} & 0.321	& 0.516	& 0.673 & 0.108	& 0.204	& 0.288 \\
\text{NN + word2vec + ppdb} & 0.360	& 0.531	& 0.730 & 0.114	& 0.213	& 0.300 \\
\text{NN + word2vec + wordnet} & 0.384	& 0.551	& 0.708 & 0.115	& 0.208	& 0.297 \\
\text{NN + word2vec + wordnet+} & 0.390	& 0.597	& \textbf{0.806} & 0.138	& 0.235	& 0.324 \\
\text{NN + random vectors} & 0.244	& 0.400	& 0.597 & -0.063	& 0.029	& 0.107 \\
\text{mode} & 0.432	& 0.554	& 0.643 & \textbf{0.420}	& \textbf{0.522}& \textbf{0.605} \\
\text{true-mode} & 0.419	& 0.551	& 0.637 & 0.379	& 0.466& 0.551 \\
\midrule
\text{\cite{herbelot2015building} (PLSR + word2vec)}		& ?	& 0.634 & ?	& ?	& 0.346 & ? \\

\bottomrule
\end{tabular}
\caption{All the presented results are averaged over 1000 runs, except for the results of~\protect\cite{herbelot2015building}) in the last row. PLSR stands for partial least squares regression, NN for nearest neighbor, ppdb for the Paraphrase Database~\protect\cite{ganitkevitch2013ppdb}. There are two ways to compute the mode: either taking the mode of the means of the 3 annotations (\textit{mode}), or the mode for all annotations (\textit{true-mode}). QMR has 3 potentially different annotations for each concept-feature pair, while AD has 3 only one annotation for each concept-feature pair: as a result, \textit{mode} and \textit{true-mode} have similar results for AD, but potentially different results for QMR. For each run, a train/test split was randomly chosen (60 training samples for AD, 400 for QMR, in order to have the same number of training samples as in~\protect\cite{herbelot2015building}'s Table 2).}\label{beta}
\end{table*}

\section{Results and discussion}

\begin{figure*}[htb!] 
\includegraphics[width=17cm]{{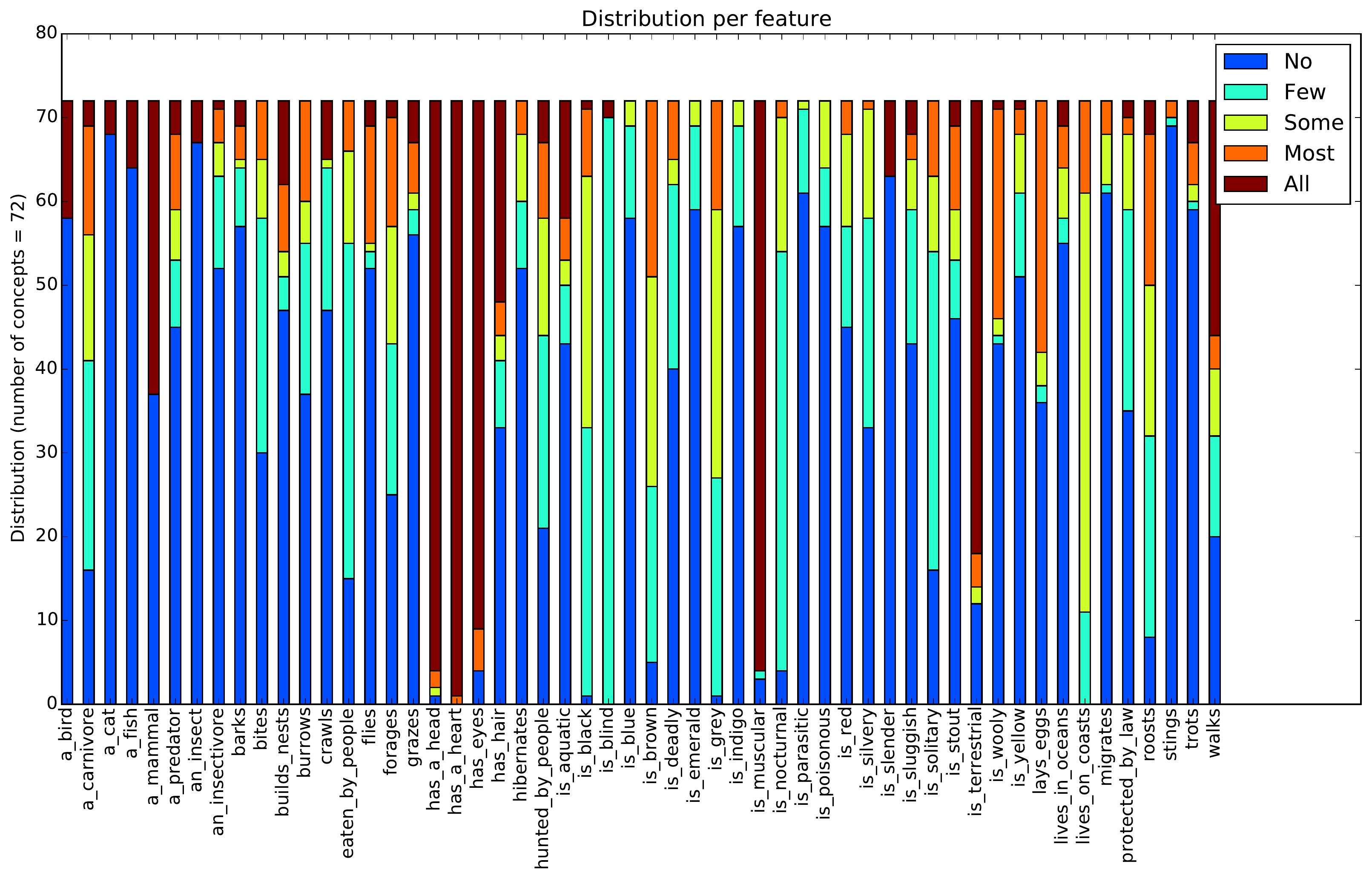}}
\caption{Stacked bars showing the distribution of quantifiers among features in the AD dataset: most features tend to have one clearly dominant quantifier. For example, the feature \textit{a\_cat} is almost always annotated with the quantifier \textit{no}.}\label{fig:ad2dbad}
\end{figure*}

\begin{figure*}[htb!]
\includegraphics[width=17cm]{{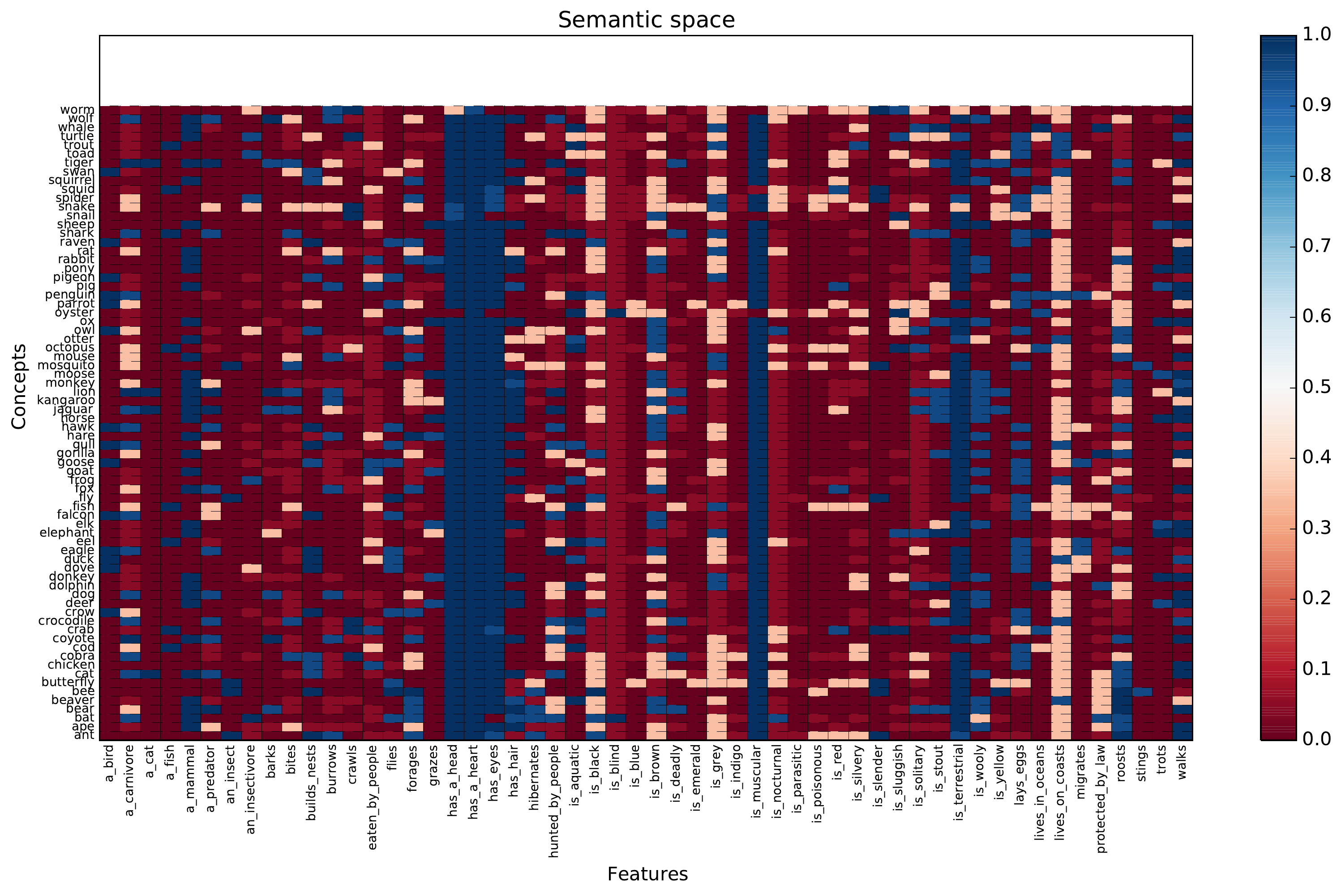}}
\caption{Heatmap showing the distribution of quantifiers among features in the AD dataset: most features tend to have one clearly dominant quantifier. The values of the heatmap are given by the following quantifier-scalar mapping: $\text{no} \mapsto 0$; $\text{few} \mapsto 0.05$; $\text{some} \mapsto 0.35$; $\text{most} \mapsto 0.95$; $\text{all} \mapsto 1$.}\label{fig:adheatmap}
\end{figure*}

Table~\ref{beta} presents the results, using the Spearman correlation as the performance metric. The experiment was coded in Python using scikit-learn~\cite{pedregosa2011scikit} and the source as well as the complete result log and the two datasets are available online\footnote{\url{https://github.com/Franck-Dernoncourt/model-theoretic}}. We could reproduce the results for the QMR dataset using PLSR and word2vec embeddings (0.346 in~\cite{herbelot2015building} vs. 0.332 in our experiments, but we could not exactly reproduce the results for the AD dataset (0.634 in~\cite{herbelot2015building} vs. 0.572 in our experiments): this discrepancy most likely results from the choice of the training set. Our experiments' results are averaged over 1000 runs, and for each run the training/test split is randomly chosen, the only constraint being having the same number of training samples as in ~\cite{herbelot2015building}. For the AD dataset, our worst run achieved 0.435, and our best run achieved 0.713, which emphasizes the lack of robustness of the results with respect to the train/test split. The variability is much lower for the QMR dataset (min: 0.244; max: 0.407), which is expected since QMR is significantly larger than AD.

\hspace{-0.0cm}Furthermore, \hspace{-0.00cm}the \textit{mode} baseline yields results that are good on the AD dataset (0.554, vs. \hspace{-0.00cm}0.634 in \cite{herbelot2015building} vs. 0.572 in our PLSR + word2vec implementation), and significantly better than all other models on the QMR dataset (0.522, vs. 0.346 in~\cite{herbelot2015building}, i.e. +51\% improvement). To get an intuition of why the mode baseline works well, Figures~\ref{fig:ad2dbad} and \ref{fig:adheatmap} show that most features tend to have one clearly dominant quantifier in the AD dataset. \hspace{-0.15cm}
A similar trend can be found in the QMR dataset.
 In the AD dataset, there are 54 features, each of them being annotated for all 73 concepts. In the QMR dataset, there are 2201 features, each of them being annotated for only $\frac{6187}{2201} \approx 2.81$ concepts on average. As a result, it is much more difficult for the PLSR to learn the mapping from word embeddings to model-theoretic vectors in the QMR dataset than in the AD dataset. This explains why the \textit{mode} baseline outperforms PLSR in the QMR dataset but not in the AD dataset.

The \textit{random vector} baseline with PLSR performs mediocrely on the AD dataset, and very poorly on the QMR dataset.
The \textit{nearest neighbor} baseline yields some competitive results on the AD dataset, but lower results on the QMR dataset. 
Lastly, using retrofitting increases the performances on both AD and QMR datasets. This is expected as applying retrofitting to word embeddings leverages relational information from semantic lexicons by encouraging linked words to have similar vector representations.

\section{Conclusion}

In this paper we have presented several baselines for mapping distributional to model-theoretic semantic spaces. The \textit{mode} baseline significantly outperforms~\cite{herbelot2015building}'s model on the QMR dataset, and yields competitive results on the AD dataset. This indicates that state-of-the-art models do not efficiently map word embeddings to model-theoretic vectors in these datasets.

\bibliography{emnlp2016}
\bibliographystyle{emnlp2016}

\end{document}